\documentclass[10pt,conference]{IEEEtran}
\let\labelindent\relax
\makeatletter
\def\BState{\State\hskip-\ALG@thistlm}
\makeatother

\usepackage[utf8]{inputenc} 
\usepackage[T1]{fontenc}    
\usepackage{times,latexsym}
\usepackage{graphicx} \graphicspath{{figures/}}
\usepackage{amsmath,amssymb,mathtools,nicefrac}
\usepackage[linesnumbered,ruled,vlined]{algorithm2e}
\usepackage{acronym}
\usepackage{enumitem}
\usepackage[breaklinks,colorlinks,bookmarks=true]{hyperref}
\usepackage{balance}
\usepackage{xspace}
\usepackage{setspace}
\usepackage[skip=3pt,font=small]{subcaption}
\usepackage[skip=3pt,font=small]{caption}
\usepackage[dvipsnames,svgnames,x11names,table]{xcolor}
\usepackage[capitalise,noabbrev,nameinlink]{cleveref}
\usepackage{booktabs,tabularx,colortbl,multirow,multicol,array,makecell,tabularray}
\usepackage{overpic,wrapfig}
\usepackage[misc]{ifsym}
\usepackage{pifont}

\usepackage[numbers]{natbib}
\usepackage{algpseudocode}
\usepackage{amsfonts}
\crefname{figure}{Fig.}{Figs.} 
\Crefname{figure}{Fig.}{Figs.}

\makeatletter
\DeclareRobustCommand\onedot{\futurelet\@let@token\@onedot}
\def\@onedot{\ifx\@let@token.\else.\null\fi\xspace}

\def\ie{\emph{i.e}\onedot}

\def\etal{\emph{et al}\onedot}
\makeatother

\acrodef{dof}[DoF]{Degree of Freedom}
\acrodef{vbts}[VBTS]{Vision-Based Tactile Sensors}
\acrodef{dom}[DOM]{Deformable Object Manipulation}

\newcommand{\sensorname}{R-Tac\xspace}

\begin{document}

    \title{PP-Tac: Paper Picking Using Tactile Feedback in Dexterous Robotic Hands}

    \author{
    \authorblockN{
    Pei Lin$^{1,2\star}$\quad
    Yuzhe Huang$^{1,3\star}$\quad
    Wanlin Li$^{1\star}$\quad
    Jianpeng Ma$^{1}$\quad
    Chenxi Xiao$^{2\dagger}$\quad
    Ziyuan Jiao$^{1\dagger}$
    }%
    \authorblockN{
    $^1$Beijing Institute for General Artificial Intelligence\quad
    $^2$ShanghaiTech University\quad
    $^3$Beihang University
    }
    \authorblockN{
    $^\star$equal contributors\quad 
    $^\dagger$corresponding authors
    }
    \href{https://peilin-666.github.io/projects/PP-Tac}{https://peilin-666.github.io/projects/PP-Tac}
    }
    

    \twocolumn[{%
    \renewcommand\twocolumn[1][]{#1}%
    \maketitle
    \vspace{-8pt}
    \begin{center}
        \centering
         \includegraphics[width=\textwidth]{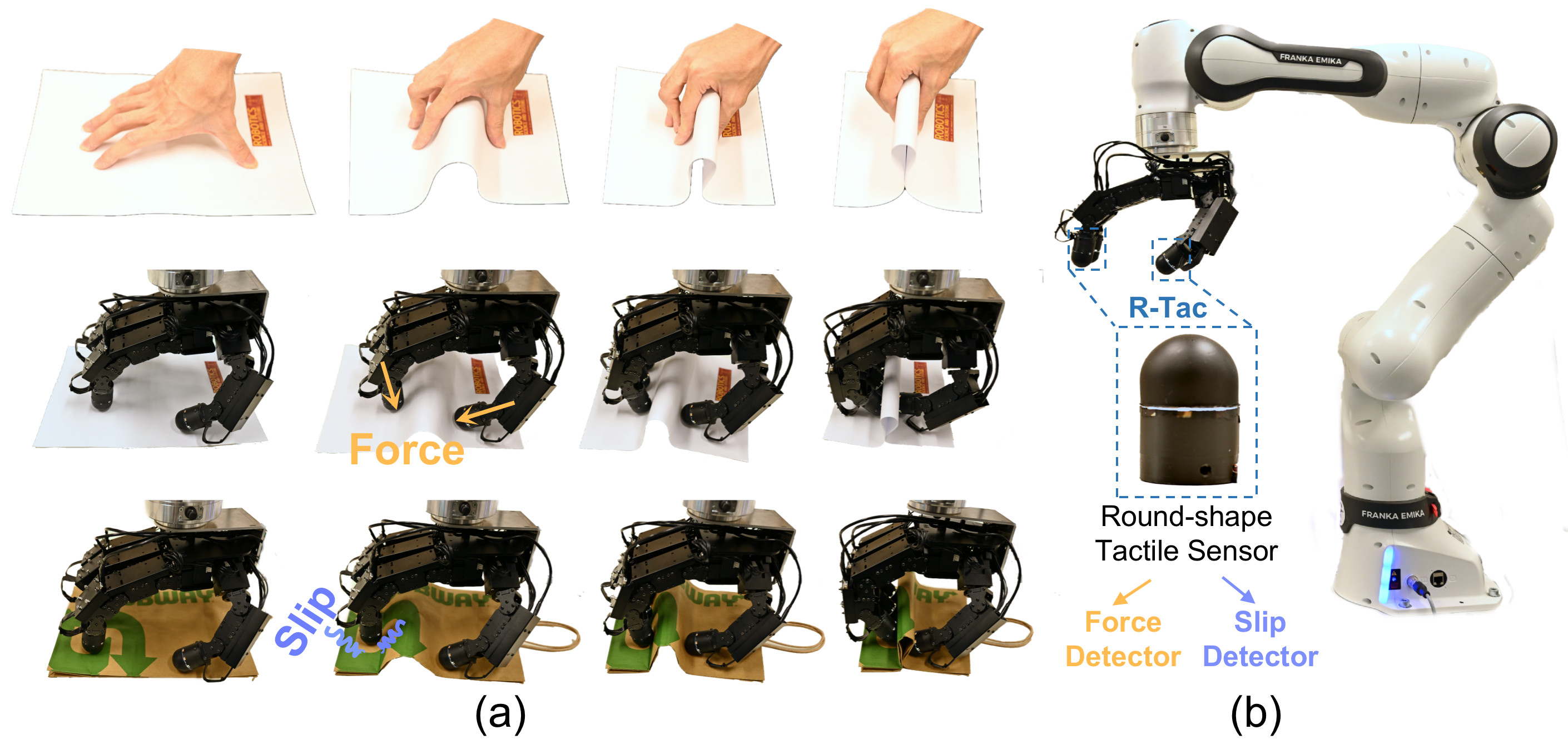}
        \captionof{figure}{\textbf{Overview of PP-Tac.}~The system leverages tactile feedback from the proposed round-shaped sensor (\sensorname), integrated into a dexterous robotic hand, to grasp thin, deformable, paper-like objects. (a) Hand motions are generated by a diffusion-based policy inspired by human strategies, such as sliding and pinching. (b) The hardware setup includes a robotic arm, a dexterous hand, and four fingertip-mounted tactile sensors that simultaneously detect force and slip events.}
        \label{fig:teaser}
    \end{center}%
    }]
    
\begin{abstract}
Robots are increasingly envisioned as human companions, assisting with everyday tasks that often involve manipulating deformable objects. 
Although recent advances in robotic hardware and embodied AI have expanded their capabilities, current systems still struggle with handling thin, flat, and deformable objects such as paper and fabric. 
This limitation arises from the lack of suitable perception techniques for robust state estimation under diverse object appearances, and the absence of planning techniques for generating appropriate grasp motions. 
To bridge these gaps, this paper introduces PP-Tac, a robotic system for picking up paper-like objects. 
PP-Tac features a multi-fingered robotic hand with high-resolution round-shaped tactile sensors \sensorname for omnidirectional tactile sensing. This hardware configuration enables real-time slip detection and online force control that mitigates slips. 
Furthermore, grasp motion generation is achieved through a trajectory synthesis pipeline, which first constructs a dataset of fingers' pinching motions. Then, a diffusion-based policy is trained to control the hand-arm robotic system. Experiments demonstrate that PP-Tac can effectively grasp paper-like objects of varying material, thickness, and stiffness, achieving an overall success rate of 87.5\%. To the best of our knowledge, this work is the first attempt to grasp paper-like objects using a tactile dexterous hand. 
\end{abstract}

\IEEEpeerreviewmaketitle

\section{Introduction}\label{sec:introduction}
Robots are increasingly popular as assistive agents in everyday life, particularly within household environments~\cite{scassellati2012robots}. These robots are designed to perform various domestic tasks, often involving the grasp of thin, deformable objects such as paper and fabric~\cite{zhu2022challenges}. For instance, clothes-folding tasks~\cite{li2015regrasping} require high dexterity and adaptability to accommodate variations in fabric size, texture, and stiffness, while document organization tasks~\cite{amigo2013general} demand picking capabilities for diverse paper types and form factors. Beyond domestic settings, handling deformable objects is essential in industrial and logistical applications, such as fabricating fabrics~\cite{billard2019trends} and packing objects using plastic bags and cardboard~\cite{dogar2011framework}.

Despite their significance, picking up paper-like objects remains challenging in robotics~\cite{zhu2022challenges}. In particular, the main challenges are three-fold: 
1) Vision systems, commonly used for manipulation, struggle to perceive contact information during interactions with deformable objects due to limited sensing modalities and occlusion, resulting in an inaccurate environment model for motion planning~\cite{li2018learning};
2) These objects are often flat in shape, lacking salient features for  contact points and thus hindering the synthesis of stable grasps ~\cite{deng2020self}.
3) The appearance of such objects exhibits high variability, as their shape undergoes continuous and unpredictable deformation during manipulation. These dynamic shape variations significantly impair the generalizability of vision-based methods.

 In contrast, humans excel at picking up paper-like objects by leveraging coordinated multi-fingered motion and tactile sensing. As shown in \cref{fig:teaser} (a), the process typically begins with establishing contact with the fingers, followed by sliding motions to deform the material and to generate a contact point for the pinched grasp. Such motion is made possible by hand's high \acp{dof}, which enables establishing multiple contact points adaptively during sliding motion. During this process, tactile sensing is also crucial as it allows humans to perceive the object’s deformation and decide the appropriate forces. These real-time adjustments ensure the successful execution of the picking-up action. 

Inspired by human strategies, this paper introduces a robotics system coined \textit{PP-Tac}: \underline{P}aper-like object \underline{P}icking using \underline{Tac}tile feedback. The PP-Tac system comprises two key hardwares: 
\textbf{A dexterous robotic hand, and the associated {hemispherical} and high-resolution \ac{vbts} {\sensorname}.} 
The fingertip-mounted tactile sensors provide real-time contact feedback during grasping operations. Featuring a spherical sensing area and a high-frame-rate monochrome camera, this design enables faster response times and simpler calibration processes compared to conventional RGB-based tactile sensors. An illustration of the system is shown in \cref{fig:teaser}(b).  In addition to the tactile sensor, this paper also presents
\textbf{A diffusion-based motion generation policy (PP-Tac policy)} that imitates human picking-up skills. The proposed method first employs efficient trajectory optimization to generate expert data replicating human sliding and pinching motions. Second, generalizability to diverse flat objects were acheived by training a diffusion policy using these trajectories, leveraging proprioceptive data and tactile feedback for adaptive control of the dexterous robotic hand.

Comprehensive real-world experiments were conducted to evaluate the PP-Tac system. PP-Tac achieved an overall success rate of 87.5\% in grasping everyday thin and deformable paper-like objects, including plastic bags, paper bags, and silk towels on flat surfaces. \cref{fig:teaser}(a) illustrates examples of our arm-hand system successfully picking up paper-like objects. The PP-Tac also demonstrates significant adaptability in picking up paper-like objects on various uneven surfaces. Additionally, an ablation study further validated the contributions of each system component, highlighting the critical role of \ac{vbts} feedback and motion generation policies in achieving robust performance.

To the best of our knowledge, this work represents the first demonstration of deformable object picking using a dexterous hand equipped with \ac{vbts}. Overall, our contributions include:  
\begin{enumerate}[label=\arabic*), leftmargin=*, noitemsep, nolistsep]
    \item We present \sensorname, a novel spherical tactile sensor designed with ease of fabrication, calibration, and scalable deployment. To demonstrate its utility, we integrate \sensorname into each fingertip of a fully actuated dexterous robotic hand, enabling real-time contact feedback during manipulation tasks.
    \item We propose a novel trajectory-optimization-based data generation framework. The proposed framework does not rely on tactile or physical simulation, which is computationally expensive, and is capable of achieving robust sim-to-real transfer.
    \item We present the \textit{PP-Tac policy}, a diffusion-based control strategy that utilizes only tactile and robot proprioceptive feedback for manipulating paper-like objects. This approach demonstrates robust generalization across diverse materials and surface properties.
    \item We provide the implementation and systematic experiments of the proposed algorithms on a real robot system. Both hardware and code for \textit{PP-Tac system} are released to support further research and community development.  
\end{enumerate}

\section{Related Work}\label{sec:related_work}

\subsection{Deformable Objects  Manipulation}

Deformable Object Manipulation (DOM) aims to handle soft objects that alter shape during interaction, which has been a long-standing challenging task in robotic research. 
Challenges mainly arised from uncertainties in perception and complex soft-body dynamics~\cite{herguedas2019survey, arriola2017multimodal,lin2023handdiffusegenerativecontrollerstwohand}. Early approaches solved such problem using visual perception for state estimation~\cite{zhu2022challenges,sanchez2018robotic}, enabling tasks like rope-handling~\cite{nair2017combining, sanchez2018robotic}, cloth-folding~\cite{sun2015accurate, li2015regrasping} and picking up paper with marker~\cite{6094742}. However, vision-based methods often fall short in solving real-world DOM problems due to varying object appearance,  object's physical property that are usually unknown in advance, visual occlusions~\cite{li2021robotic, boroushaki2021robotic}, and variable lighting conditions~\cite{wu2019illumination, krawez2021real}. 
These challenges hinder the development of scalable vision-based DOM solutions for diverse environments.

Tactile sensing, particularly \acf{vbts}, has demonstrated significant potential for solving DOM tasks~\cite{zhu2022challenges}.  Leveraging their high-resolution tactile feedback, \ac{vbts} has demonstrated high performance in object shape reconstruction~\cite{ota2023tactile,do2022densetact,lin20239dtactcompact,ward2018tactip}, localization~\cite{kerr2023selfsupervised,li2014localization,chaudhury2022using}, and slip detection~\cite{taylor2022gelslim,dong2019maintaining}. Prior work has explored \ac{vbts} for deformable object manipulation~\cite{she2020cable}, but existing implementations rely on gripper-mounted sensors, which lack the dexterity of multi-fingered hands due to limited \ac{dof}. Our experiments reveal that gripper-based approaches struggle with thin deformable objects, and lack sufficient adapability for objects placed on non-flat surfaces, which highlights the need for integrating dexterous robotic hands with \ac{vbts} for robust manipulation~\cite{khandate2023sampling}.

\begin{figure*}[h!]
    \centering
    \includegraphics[width=\linewidth]{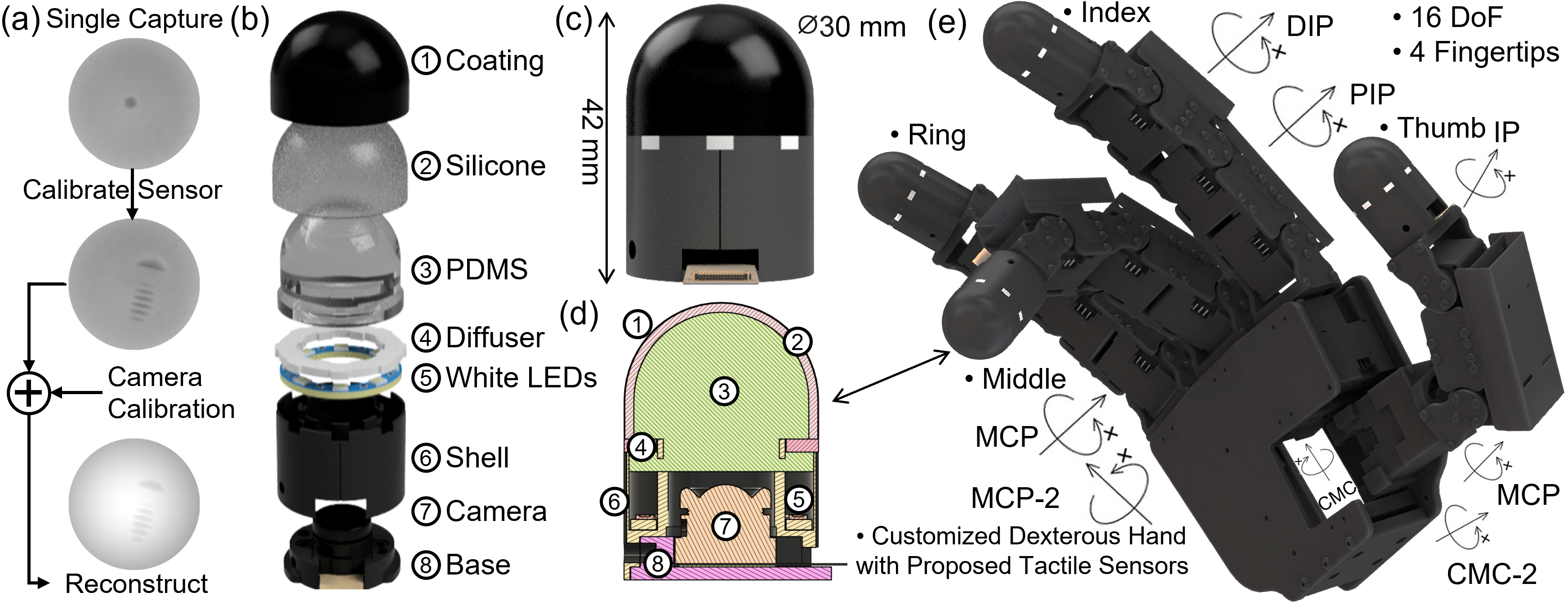}
    \caption{\textbf{The hardware design of the \sensorname and its integration into the four-fingered dexterous robotic hand system.} (a) illustrates the pipeline of depth reconstruction. (b) illustrates the exploded view of the sensor, detailing each component. (c) shows the dimensions of the sensor. (d) shows the schematic design. (e) illustrates the robotic hand equipped with four sensors as its distal links.}
    \label{fig:hardware}
\end{figure*}

\subsection{Dexterous Robotic Hand with Tactile Sensing}
Current dexterous hands are often equipped with tactile sensors. Commonly used tactile sensors typically incorporate mechanisms such as capacitive~\cite{liu2024material}, piezoresistive~\cite{kaboli2015hand}, or magnetic-based~\cite{funabashi2020stable} technologies. These designs can be fabricated in various shapes and sizes, allowing them to conform to the form factor of different robotic fingers. However, the sensing principles behind these technologies limit their spatial resolution and robustness under varying environmental conditions. To enhance sensing quality, recent work has devoted efforts to develop various \ac{vbts} sensors, especially with curved elastomer surfaces~\cite{do2022densetact, tippur2024rainbowsight, lambeta2024digitizing, andrussow2023minsight, azulay2023allsight}. However, most of these VBTS sensors are not yet commercially available, and remain challenging to deploy at scale in hand. This is mainly due to the challenge arised from sensor's calibration, as the illumination from RGB chromatic light sources results in uneven light intensity distributions on curved elastomer surface, necessitating extensive data collection for calibration. During such data collection, these sensors~\cite{do2022densetact, tippur2024rainbowsight, lambeta2024digitizing, andrussow2023minsight} often require specialized test beds (\textit{e.g.}, fabricated using CNC machines) to collect large datasets, increasing the calibration complexity. Moreover, transmitting realtime chromatic video streams imposes higher bandwidth requirements, which can limit the overall frame rate in large-scale deployments.
To address the above issues, we propose \sensorname that is structurally simple to fabricate, compact, and easy to  calibrate.

Current robotic hands equipped with \ac{vbts} have been used in grasping and in-hand orientation tasks. For instance, Do~\etal uses DenseTact~\cite{do2022densetact} attached to an Allegro Hand, to grasp and manipulate small screws~\cite{do2023inter}. Qi~\etal integrates fingertip \ac{vbts}~\cite{qi2023general} and DIGIT~\cite{suresh2024neuralfeels} on an Allegro Hand to rotate objects in hand.
To the best of our knowledge, existing research has not yet explored \ac{vbts}-equipped dexterous hands for manipulating thin, deformable objects such as paper sheets.

\section{Hardware Design}\label{sec:hardware}
To provide sufficient dexterity to address the challenges of paper-picking tasks, we designed and fabricated a set of round-shape \ac{vbts}---\sensorname, which are integrated into Allegro Hand~\cite{AllegroHand} through customization.

\subsection{Fingertip-shaped Tactile Sensing}

The design of \sensorname is guided by five key principles to ensure effective manipulation:

\begin{itemize}[leftmargin=*, noitemsep, nolistsep]
    \item \textbf{Round shape:} The hemispherical design enables omnidirectional tactile perception.
    \item \textbf{High resolution:} High resolution enables accurate depth reconstruction and slip detection during picking-up.
    \item \textbf{Convenient to fabricate \& low-cost:} The components of the tactile sensor are either off-the-shelf or easy to fabricate, with a cost of around \$60.
    \item \textbf{Efficient calibration:} The monochrome sensing principle simplifies lighting control and reduces manual effort for calibration, making it particularly suitable for large-scale deployment on multi-fingered robotic hands.
    \item \textbf{Efficient data transmission:} The monochrome camera produces lightweight data per frame, facilitating high-speed data transmission between systems.
\end{itemize}

Based on these 5 principles, the sensor design and its integration into the dexterous hand is illustrated in \cref{fig:hardware}. Next, we detail each component and the calibration process.

\subsubsection{Contact and Illumination Module}
The core of the sensor is a contact module (elastomer) with a uniformly illuminated, deformable sensitive surface that maintains structural rigidity during contact. Inspired by the monochrome sensing principle~\cite{lin2023dtact}, where intensity changes indicate deformation, we developed a hemispherical structure comprising a white LED ring, a stiff transparent internal skeleton, a soft semitransparent perception layer, and a thin opaque protective layer that achieves the desired optical characteristics.

The LED ring (LUXEON 2835 4000K SMD LED) and a diffuser (double-sided frosted diffuser sheet) are first installed within the sensor shell. The skeleton is then manufactured from PDMS (Dow Corning
Sylgard 184 with Shore hardness 50 A) using a two-piece molding technique. The mixture (base: catalyst = 10: 1) is degassed and poured into the mold, and cured for 24 hours at room temperature. The perception layer is then manufactured similarly, using semitransparent silicone (Smooth-On Ecoflex with Shore hardness 00-10), and the layer is peeled off after 4 hours. Note that the measured depth range relies on the thickness of this layer, which is set to 2 mm. Finally, a silicone coating (Smooth-On Psycho Paint) is airbrushed onto the perception layer to form the opaque protective layer. The entire manufacturing process takes within 3 days, facilitating large-scale deployment.

\subsubsection{Camera Module}
A micro black-and-white CMOS camera (OV9281) with a wide $160^{\circ}$ lens is used to capture the light intensity data. The camera operates up to 120Hz with a resolution of $640\times480$ and a latency of approximately 100ms.

\subsubsection{Calibration}
The uniform optical properties of the elastomer and illumination module (with a capture standard deviation as low as 6) enable the 3D geometry of the round shape sensor to be computed from single-channel pixel intensity in simply two steps using only 30 captures, without the need for a CNC machine. First, given the known intrinsic parameters \(K\), camera calibration is performed using 29 captures in a 3D-printed indentation-based setup to estimate the extrinsic parameters of rotation matrix \(A\) and translation vector \(b\), as well as the sensor surface reference projection \(D\). Next, the depth mapping function $M$ is calibrated by capturing a single image of a ball of known size pressed onto the sensor~\cite{lin2023dtact}. The complete mapping function from the pixel coordinates \((u,v)\) to the sensor coordinates \((x,y,z)\) can be expressed as:


\begin{align}
\resizebox{.9\linewidth}{!}{$
    \begin{bmatrix}
     x\\
     y\\
     z
    \end{bmatrix} = A^{-1}
    \begin{pmatrix}
    (D(u,v)-M(I_{\Delta}(u,v)))
    K^{-1}
    \begin{bmatrix}
    u \\
    v \\
    1
    \end{bmatrix} - b
    \end{pmatrix} ,
    $}
\end{align}
which transforms grayscale intensity images to a depth map expressed in the sensor coordinates. A detailed explanation of camera calibration is provided in \Cref{appendix:cameracali}. Reconstruction results and qualitative analysis are presented in \Cref{exp:depth_recon}.

\subsubsection{Contact Force Estimation \& Slip Detection}\label{subsec:force&slip}
\begin{figure}[t]
    \centering
    \includegraphics[width=\linewidth]{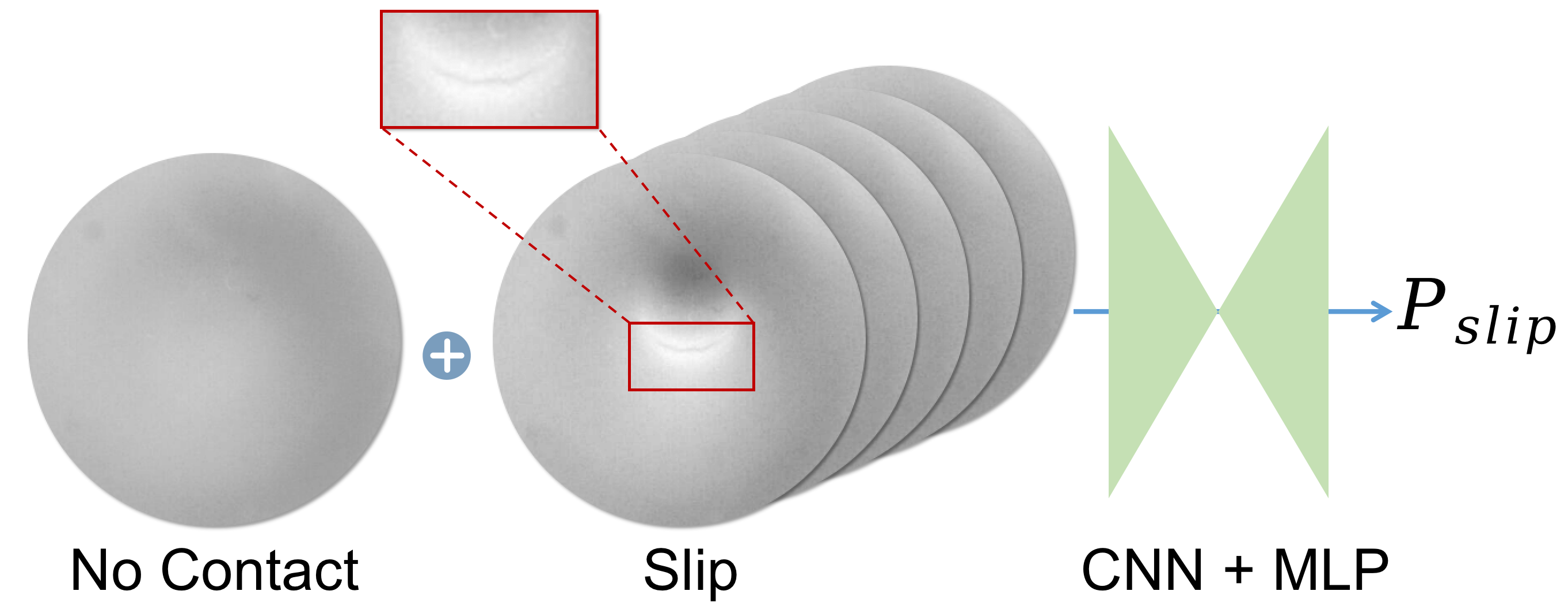}
    \caption{\textbf{Slip detection.} The left tactile image shows no contact, while the middle tactile image highlights wrinkle features during slip. The network computes the probability of slipping $P_{slip}$ using the no-contact tactile image and five most recent tactile images.}
    \label{fig:slip}
    
\end{figure}

Our sensors are capable of detecting both contact forces and slip events. The contact force, modeled by elasticity theory, is proportional to the deformation depth as a linear function. The slip detection module is as follows:

\begin{itemize}
    \item \textbf{Detection Model:} 
     As illustrated in Fig.~\ref{fig:slip}, when the slip occurs, distinct wrinkles become visible in the sensor's imaging. We apply a lightweight neural network architecture consisting of CNN (convolutional neural network) and MLP (multilayer perceptron) to detect the slip. 
     The network processes a temporal sequence of the preceding five frames with a non-contact frame as input. CNN extracts the feature per image and then the feature maps are concatenated together to estimate the slip probability \( P_{\text{slip}} \) through a MLP.

    \item \textbf{Training:} 
     To train the network, we collected approximately 20 minutes of tactile data from four sensors. The dataset comprises 40\% slip samples and 60\% non-slip samples, with each frame manually annotated. We choose binary cross-entropy as our loss function. 

    \item \textbf{Inference:} 
     In inference, it is necessary to set a threshold for \( P_{\text{slip}} \) for positive detection. Through manual adjustments, empirical results demonstrate that a threshold of 0.75 yields an optimal trade-off between sensitivity and accuracy, achieving a slip detection accuracy of 86\%. 
\end{itemize}

\subsection{Robotic Hand System}
We integrated the proposed \sensorname sensors into a fully actuated dexterous robotic hand. These tactile sensors are mounted at the distal end of each fingertip, facilitating contact characterization in the following paper-picking tasks. We designed and fabricated the robotic hand featuring 16 controllable \acp{dof}, including the DIP, PIP, and MCP, MCP-2 joints for the index, middle, and ring fingers, as well as the CMC, CMC-2, MCP, and IP joints for the thumb. The robotic hand is driven by Dynamixel XC330-M288-T motors, which are all multiplexed through a U2D2 Hub. For each tactile sensor, it communicates with the PC via a USB interface. The entire assembly is mounted on a Franka Research 3, a 7-DoF robotic arm, which communicates with the PC via a high-speed Ethernet connection.

\begin{figure}[t!]
    \centering
    \includegraphics[width=.8\linewidth]{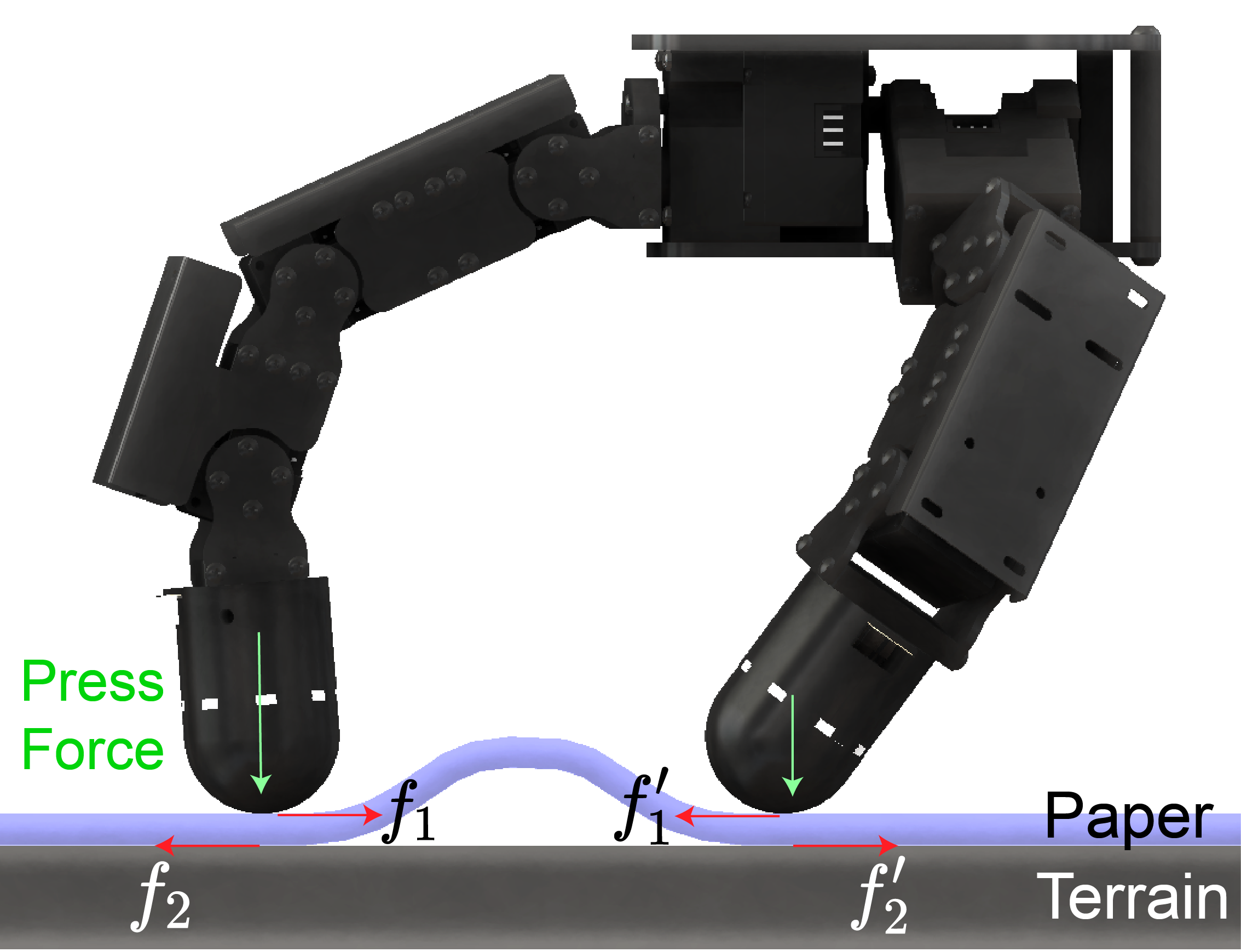}
    \caption{\textbf{Force analysis during grasping flat objects.} The grasping process is made possible by the following forces: 
    (1) the contact normal force exerted by the sensor on the object.
    (2) the static friction force (\(f_1,f_1'\)) between fingers and the object, (3) a dynamic friction force (\(f_2,f_2'\)) between the object and the terrain. When the static friction (\(f_{1},f_{1}'\)) exceeds the critical buckling resistance of the paper, the sheet deforms, creating a stable pinch region that facilitates successful grasping.
 }                    
    \label{fig:problem}
\end{figure}
\section{Paper-like Object Picking Problem Statement}\label{sec:problem_statement}

Next, we aim to address the challenge of grasping thin, deformable paper-like objects from flat surfaces. This appears as a commonly seen scenario in everyday tasks, such as organizing scattered document pages or retrieving napkins from dining plates. Although creases or irregularities in the material can sometimes provide grasping points, a particularly challenging scenario arises when the object is extremely flat and lacks discernible edges or salient grasping features. This research introduces a novel approach to tackle this paper-picking problem that was previously unexplored.

Motivated by the human strategy for grasping flat objects, our work is based on a biomimetic grasping pose optimized for paper picking, as illustrated in \cref{fig:problem}. By applying sufficient inward force, the robotic fingers can induce buckling of the material against the supporting surface. This buckling effect dynamically generates a pinchable region, enabling subsequent grasp execution.

During buckling, the distance between contact points beneath the fingers decreases. When this reduction rate matches the fingertips’ closure speed (\ie, no relative motion between fingertips and material), two frictional forces govern the system: static friction (\(f_1, f_1'\)) between the fingers and material, and dynamic friction (\(f_2, f_2'\)) between the material and the supporting surface. Their magnitudes depend on the applied normal force and the respective coefficients of friction.

In particular, the above analysis assumes that the static friction between robotic fingers and the material exceeds both the maximum static friction at the material-terrain interface and the critical buckling resistance of the material. This framework can also be extended to scenarios with uneven supporting surfaces. Without loss of generality, we assume that height variations in the terrain are less than 3 cm. 

One challenge is determining the control inputs for all joints and the wrist pose. Intuitively, this resembles human grasping behavior: when picking up a sheet of paper from a flat surface, the wrist must first elevate and then lower to establish stable contact. However, a finger-wrist coupling issue arises: the motion of one finger requires a specific wrist state, which in turn affects the movement of other fingers. In practice, our approach solved this problem by adopting a learning-based policy rather than a model-based optimization paradigm. This is due to its superior efficiency in deployment, as we found model-based optimization is too computationally expensive to adapt for online execution.

\section{Policy Learning for Paper-Picking}\label{method}
Manipulating paper-like objects with visual perception remains challenging due to difficulties in detecting thickness and textural variability. To address this, we propose a vision-independent tactile-based approach. The core idea leverages tactile feedback to maintain contact conditions (as defined in~\cref{sec:problem_statement}), facilitating the creation of a buckling region for successful grasping. We implement this through the \emph{PP-Tac policy}, developed in two stages: 1) Trajectory Optimization: Generate a dataset of grasping motions using trajectory optimization. 2) Diffusion Policy Training: Train a policy on this dataset to infer motions from tactile feedback and proprioceptive states, ensuring generalization to real-world robotic systems.

\subsection{Grasp Motion Dataset Synthesis}\label{method:establish_contact}

    \begin{figure}[t]
        \centering
        \includegraphics[width=\linewidth]{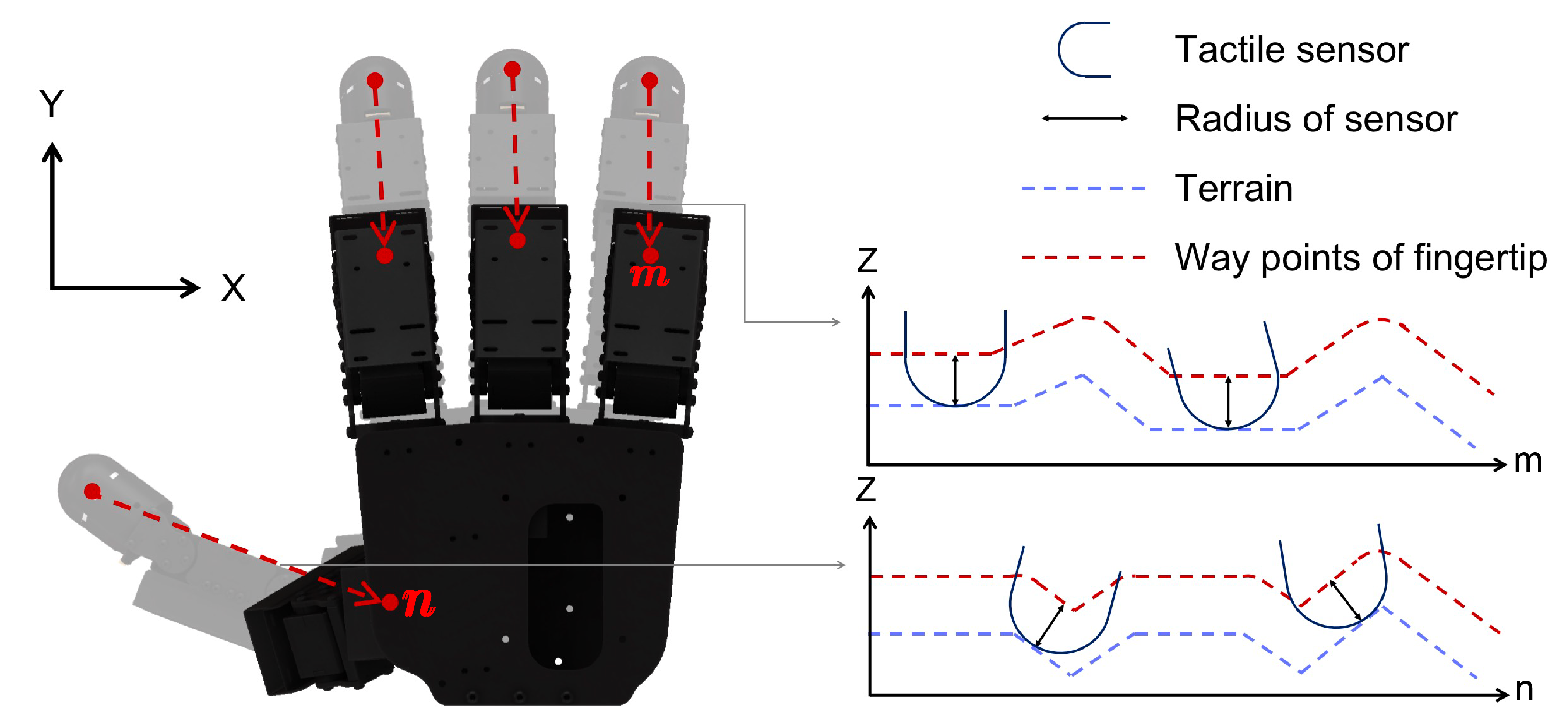}
        \caption{\textbf{Fingertip trajectories from data synthesis.} Trajectories ensure fingertip sliding along the terrain surface. Adjusting the distance between waypoints and terrain affects sensor deformation. The right figure projects trajectories of two fingers onto the \(m\)-\(z\) and \(n\)-\(z\) planes, where \(m\) and \(n\) are straight-line projections of fingertip trajectories on the palm-aligned \(x\)-\(y\) plane, and the \(z\)-axis extends outward from the hand.}

        \label{fig:DataSynthesis}
    \end{figure}

We synthesize grasping motions through trajectory optimization in simulation, avoiding the need for complex teleoperation devices. While reinforcement learning (RL) offers an alternative, it requires soft-body simulation to model deformable object dynamics and \ac{vbts} elastomer behavior, often necessitating additional real-to-sim procedures for fidelity. In contrast, our approach uses rigid-body dynamics and transfers directly to real robots, as validated experimentally. The grasping process begins by establishing contact between the fingertips and the object’s surface (see \cref{appendix:InitializeState} for implementation details). Once contact is achieved, the fingers gradually close to pinch the object. Each finger follows an independent trajectory on the object’s surface, and concurrently exerts target normal forces (\cref{fig:problem,fig:DataSynthesis}).

To generate diverse fingertip trajectories, we first generated randomized terrain profiles and pre-recorded a grasping motion sequence. The grasping motion sequences were obtained through teleoperation, capturing natural and usual grasping motion. As illustrated in \cref{fig:problem,fig:DataSynthesis}, we then extracted the (x,y) coordinates of the fingertip trajectories from the pre-recorded motion to serve as target positions. The corresponding z-coordinates were obtained by projecting these (x,y) points onto the terrain surface and sampling the z-value at each location. 

To account for varying material properties, sliding trajectories are adjusted based on fingertip contact forces \(F\). This is achieved by synthesizing motions that deform the tactile sensor's elastomer layer to different extents. The deformation reading \(\pmb{d}_{tac}\) is proportional to the applied pressure, governed by the elastomers Young's modulus. Thus, contact force \(F\) is modulated by controlling \(\pmb{d}_{tac}\) via position control. Notably, the exact relationship between \(\pmb{d}_{tac}\) and \(F\) is not explicitly modeled, as precise force values are unnecessary for the algorithm. Our approach leverages rigid-body dynamics to control contact forces efficiently, avoiding complex deformable dynamics calculations. 
By adjusting the distance between the finger joint and the terrain, we can obtain trajectories with varying degrees of deformation. For example, when the distance between the finger joint and the terrain is equal to the sensor's radius (\Cref{fig:DataSynthesis}), the finger just makes contact with the terrain and \(\pmb{d}_{tac}\) just equals to 0. Finally, we get the target trajectories of 4 fingertips, with their positions denoted as $\pmb{ee}_{target}$.

Given $\pmb{ee}_{target}$, all of the finger joint angles and arm poses are solved through the following optimization problem:
\begin{align}
    \hat{\pmb{\gamma}} &= \arg\min_{\pmb{\gamma}} \left( L_{\pmb{ee}} + L_{\Delta} + L_{\pmb{R},\pmb{p}_{wrist}} \right), \\
    L_{\pmb{ee}} &= w_{\pmb{ee}}\  \mathbf{ MSE}( \mathbf{fk}(\pmb{\gamma}),  \pmb{ee}_{target}), \\
    L_{\Delta} &= w_{\Delta} \ \mathbf{ MSE}\left( \bar{\pmb{\gamma}},  \pmb{\gamma}\right), \\
    L_{\pmb{R},\pmb{p}_{wrist}} &= w_{\pmb{R},\pmb{p}_{wrist}}\  \mathbf{ MSE}\left((\pmb{\bar{R}},\pmb{\bar{p}}_{wrist}),\right. \nonumber \\
    &\quad \ \ \ \ \ \ \ \ \ \ \ \ \ \  \ \ \ \ \ \ \  \left. (\pmb{R},\pmb{p}_{wrist})\right),
\end{align}
where \(\pmb{\gamma}\) is the optimization variables consisting of \(N_{data}\) frames' hand joint angles \(\pmb{q}\), wrist (end effector of arm) rotation \(\pmb{R}\) and wrist translation along the \(z\)-axis in world coordinates \(\pmb{p}_{wrist}\). \(N_{data}\) is the sequence length. The forward kinematics \(\mathbf{fk}\) computes the four fingertips' trajectories by giving \(\pmb{\gamma}\). \(\mathbf{MSE}\) denotes mean squared error. \(L_{\pmb{ee}}\) can minimize the error between the fingertip positions and their targets, while \(L_{\Delta}\) regularizes the motion to remain close to the initial pose. Additionally, \(L_{\pmb{R},\pmb{p}_{wrist}}\) minimizes wrist movement, ensuring the arm stays within its workspace. In practice, we choose SGD as our optimizer. After we filtering out the sequence with collision, we generated a dataset of 500,000 grasp samples, each comprising a sequence of \(N_{data} = 100\) frames.

\subsection{PP-Tac Policy}\label{subsec:DiffusionPolicy}

\begin{figure*}[t!]
    \centering
    \includegraphics[width=\linewidth]{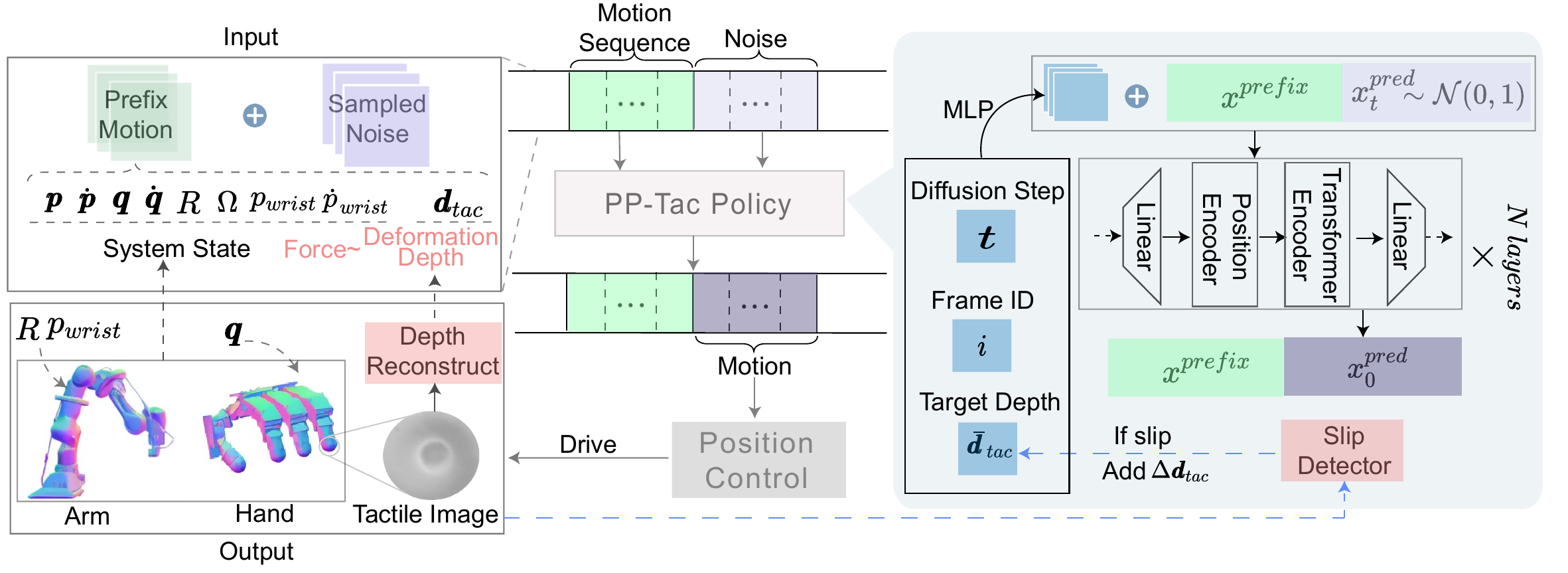}
    \caption{\textbf{Inference pipeline of the proposed PP-Tac policy.} Conditioned on robot proprioception and the target force that needs to be exerted, PP-Tac can infer the action of the next steps. If slip is detected between the finger and the flat object underneath, an incremental amount of force will be exerted by the finger. }
    \label{fig:pipeline}
\end{figure*}

Once the dataset is prepared, we employ a diffusion policy to jointly control the hand and arm, enabling adaptation to varying terrain shapes and contact force conditions. We adopt a Denoising Diffusion Probabilistic Model (DDPM) framework~\cite{DDPM,ho2022imagenvideohighdefinition,chi2023diffusion,song2020score}, which predicts future actions (\( N_{pred} \) steps of \( x^{pred} \)) conditioned on historical states (\( N_{prefix} \) steps of \( x^{prefix} \)). In each frame, the state variables include:  
\[
(\pmb{p},\dot{\pmb{p}},\pmb{q},\dot{\pmb{q}},R,\Omega ,{p_{wrist}},\dot{p}_{wrist}, \pmb{d}_{tac})
\]

where \( \pmb{p} \in \mathbb{R}^{17 \times 3} \) is hand joints' position in world coordinate,  \(\dot{\pmb{p}} \in \mathbb{R}^{17 \times 3} \) is the linear velocity of the hand joints relative to each parent frame, \( \pmb{q} \in \mathbb{R}^{16
} \) is the rotation angle of controllable hand joints,  \( \pmb{\dot{q}} \in \mathbb{R}^{16} \) is the angular velocity of controllable hand joints, \( R \in \mathbb{R}^6 \) is 6D rotation (represented as two-row vectors of rotational martix, which is from \cite{Zhou_2019_CVPR}) of wrist(end effector of arm), \(\Omega \in \mathbb{R}^6 \) represents the angular velocity of wrist rotation, \(p_{wrist} \in \mathbb{R} \) is the wrist's height along arm's \( z \)-axis, \(\dot{p}_{wrist} \in \mathbb{R}\) is the linear velocity of \(p_{wrist}\), \( \pmb{d}_{tac} \in \mathbb{R}^4 \) represents the deformation depth readings from four fingertip tactile sensors. \cref{tab:symbols} summarizes the notations used in this paper. The total state dimension is \( \mathcal{D}  = 152\). Such a over-parameterized input allows the network to extract more robust and expressive latent features for the diffusion policy.

The pipeline is illustrated in \Cref{fig:pipeline}. \Cref{fig:pipeline} (right) illustrates a single denoising diffusion step.
We apply an encoder-only transformer to predict future robot motion \( x_0^{pred} \) given prefix motion \( x^{prefix} \), diffused future motion \( x_t^{pred} \), diffusion step \( t \), current frame index \( i \), and target deformation depth \( \pmb{\bar{d}}_{tac} \). The input sequence is encoded into a latent vector of dimension \( \mathbb{R}^{(1 + N_{prefix} + N_{pred}) \times \mathcal{D}} \), comprising: 1) A latent vector of \( \mathcal{D}  \)-dimensional features representing \( t \), \( i \), and \( \pmb{\bar{d}}_{tac} \) extracted using a three 3-layer MLP network. 2) \( N_{prefix} \times \mathcal{D}  \) dimensions corresponding to the prefix states of $N_{prefix}$ time steps. 3) \( N_{pred} \times \mathcal{D}  \) dimensions for the predicted states of $N_{pred}$ time steps. 
Instead of predicting $\epsilon_t$ (formulated by \cite{ho2022imagenvideohighdefinition}), we follow \cite{tevet2023human} to predict the state sequence itself $\hat{x}_0^{pred}$.
Predicting $\hat{x}_0^{pred}$ is found to produce better results for the state sequence which contains motion data, and enables us to apply a target loss for each denoising step as following:
  
\begin{align}
    L &= \lVert \hat{x}_0^{pred} - x_0^{pred} \rVert_2^2 + \lambda_{consist} L_{consist}, \\
    L_{consist} &= \lVert \mathbf{fk}(\pmb{q}_0^{pred}) - \pmb{p}_0^{pred} \rVert_2^2
\end{align}
where \( L_{consist} \) enforces consistency between joint angles and positions, and \( \lambda_{consist} \) is a weight hyper-parameter.

During inference, we set \(t=1000\) and the diffused \( x_{1000}^{pred} \sim \mathcal{N}(0, I) \) and iteratively denoise it to produce \( x_0^{pred} \). To ensure real-time performance, we reduce denoising steps to 10 and set \( N_{pred} = N_{prefix} = 5 \), achieving motion generation in 11 ms on an RTX4090 GPU. The predicted \( \pmb{q} \) controls the hand, while \( R \) and \( p_{wrist} \) control the arm.  

During grasping, preventing slip between the object and the fingertips is essential to maximize material deformation. To achieve this, a fingertip contact force controller is introduced, which adjusts the fingertip's deformation depth \( \pmb{d}_{tac} \). If slip is detected by the tactile sensors, we increase the desired deformation depth by a small increment \( \Delta \pmb{d}_{tac} \). 

To deploy diffusion policy to real robots, we also need to tackle the domain gap between the real world and simulation. This is achieved by introducing four distinct ways to incorporate disturbances into \(x^{prefix}\) during training.

\begin{itemize}
\item Add random Gaussian noise to $\pmb{\gamma}$ to simulate various control errors that may occur in real-world situations.

\item Add Gaussian noise to the first frame and gradually amplify it in subsequent frames, simulating the fingers moving across a rising or descending terrain.

\item Randomly choose from 2 to $N_{prefix}$ temporal consistent frames to be static, simulating fingers getting stuck due to excessive pressure on complex terrain. And $\pmb{d}_{tac}$ is set to its maximum threshold. The reason for adding the index of the frame into the input is also to avoid issues caused by the fingers getting stuck.
\end{itemize}

\section{Experiments}\label{sec:experiments}

In this section, we present comprehensive experiments to evaluate our proposed PP-Tac pipeline. First, we detail the implementation of our algorithm (\Cref{exp: implementation details}).
Next, we show the quantitative and qualitative results of the depth reconstruction of our \ac{vbts} (\Cref{exp:depth_recon}).
Then, we perform systematic comparisons of our system on different flat materials and supporting terrains (\Cref{sec:comparison}).
We also compare our system with various manipulators to highlight its advantages and limitations (\Cref{sec:exp:other_conf}). 
Last, ablation studies are conducted to examine the influence of parameters and the necessary training steps (\Cref{sec:ablation}).

\begin{figure}[th!]
    \centering
    \includegraphics[width=\linewidth]{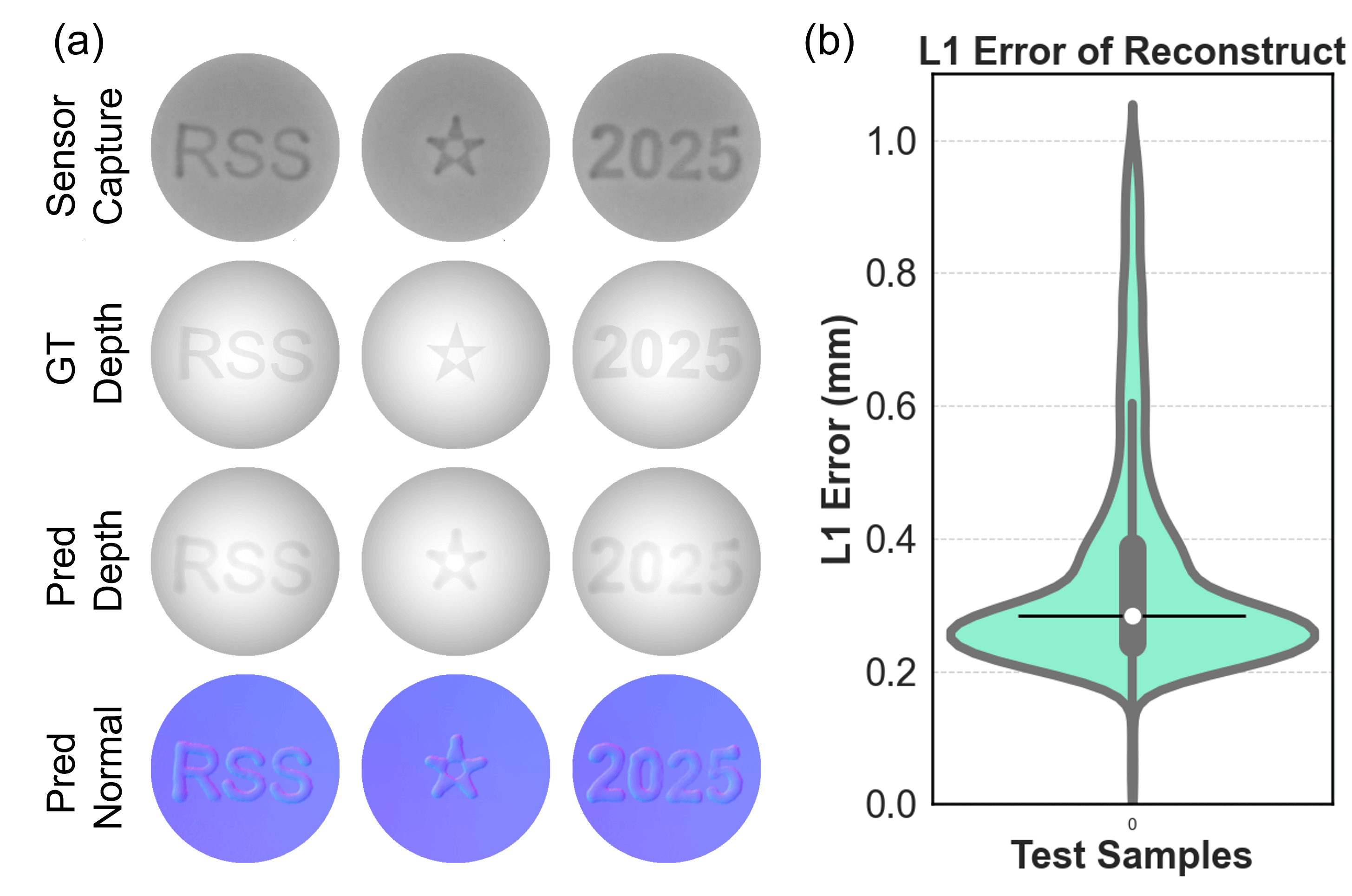}
    \caption{\textbf{Reconstruction results.} (a) Gallery of reconstructed depth and normal maps from tactile images. (b) Depth reconstruction error of indentation test.}
    \label{fig:reconstruction}
    
\end{figure}
    \begin{figure*}[ht]
        \centering
        \includegraphics[width=\linewidth]{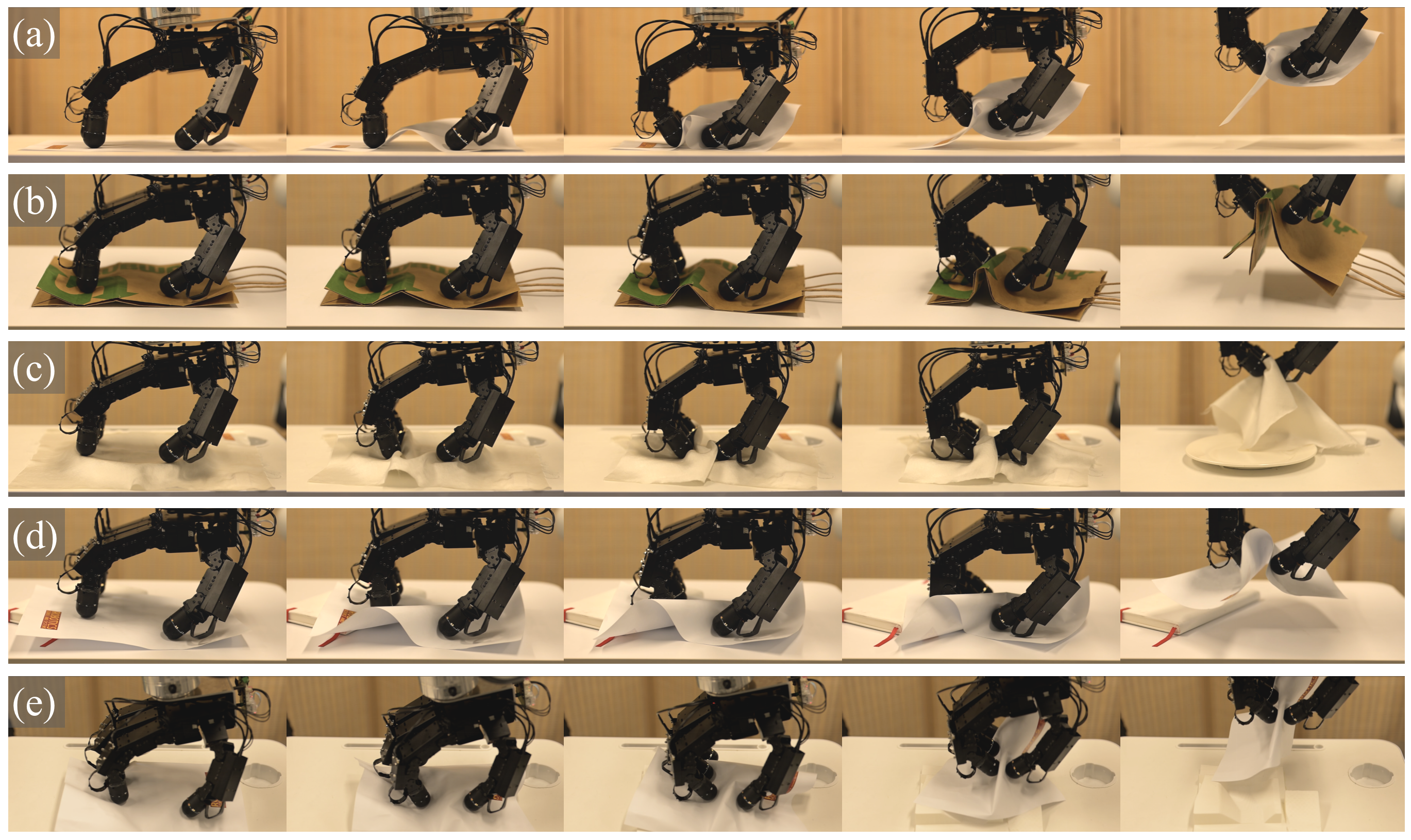}
        \caption{\textbf{Gallery of Grasping Different Objects in Real-World Evaluations.} This figure demonstrates successful grasps of five flat objects on four different types of terrains, highlighting the effectiveness of our hardware and the PP-Tac algorithm. (a) A paper sheet on a flat desktop. (b) A stiff kraft paper bag on a flat desktop. (c) A soft napkin on a plate. (d) A paper sheet on a randomly arranged book. (e) Paper sheet on a random terrain.
        These evaluations showcase the robustness and adaptability of our approach.}
        \label{fig:gallery}
    \end{figure*}

    \begin{figure*}[ht]
        \centering
        \includegraphics[width=\linewidth]{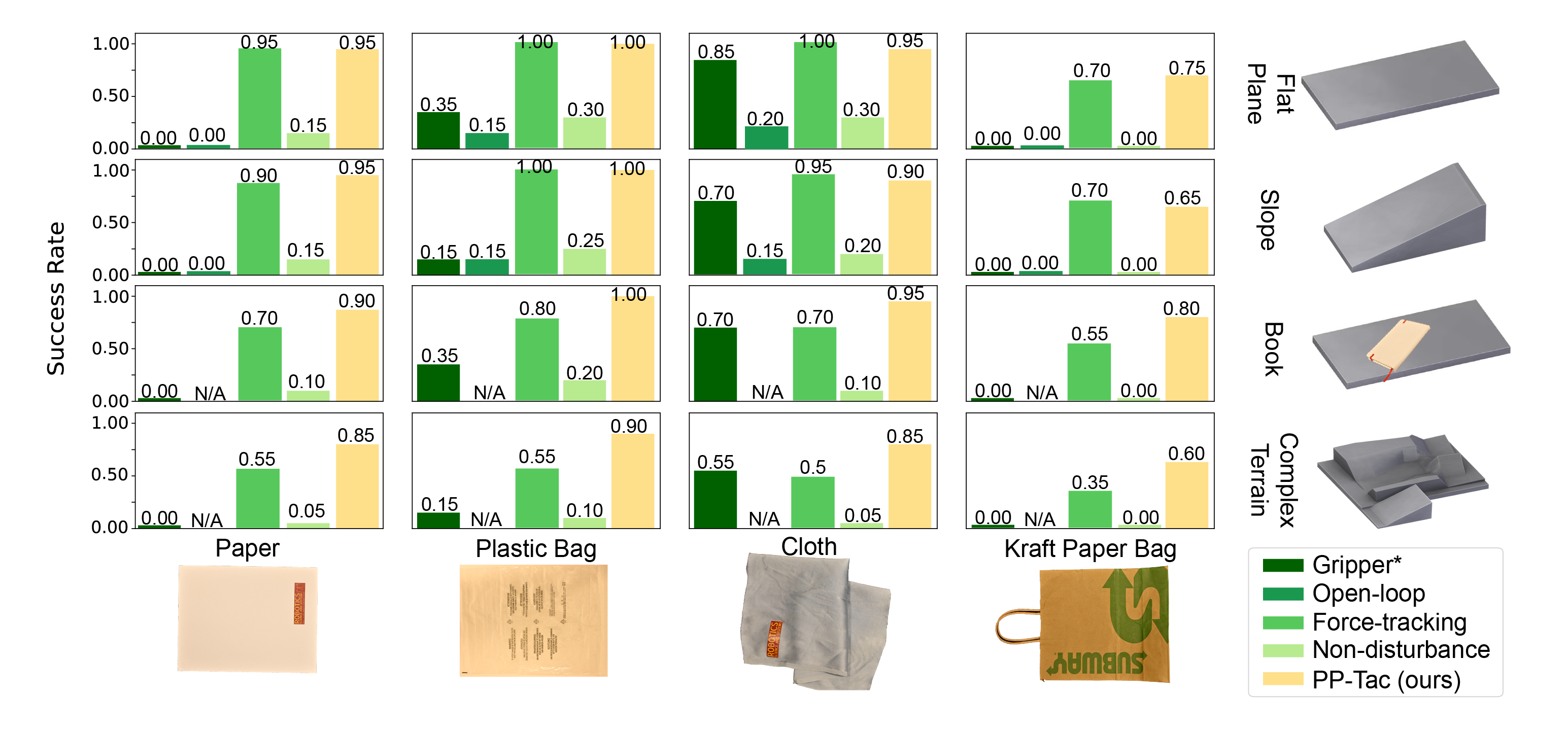}
        \caption{\textbf{Experimental Results.} Evaluations were conducted to quantify the success rate of grasping four different flat objects (paper, plastic bag, cloth, and paper bag) across four terrain setups (plane, slope, book placement, and randomly generated complex terrain). Baseline conditions included: (1) Gripper\(^*\): grasp using a bi-finger gripper controlled by teleoperation; (2) Open-loop: baseline combines the PP-Tac-derived hand trajectory with compliant finger control via tactile feedback; (3)``Model based force tracking": combines the PP-Tac-derived hand trajectory with compliant finger control via tactile feedback; (4) Non-disturbance: grasp using our dexterous hand with tactile sensors, where the diffusion policy was trained without domain randomization disturbances; and (5) PP-Tac(ours): grasp using our full PP-Tac pipeline. Each condition was repeated 20 times. Note that open-loop grasp control is not feasible on uncertain terrains, and these cases are labeled as `N/A'.}
        \label{fig:comparison}
    \end{figure*}

\subsection{Implementation Details} \label{exp: implementation details}
For reproducibility, we provide the implementation details of the PP-Tac algorithm. Our diffusion policy is implemented as a four-layer Transformer encoder with a latent dimension of 512 and four attention heads. We split each synthesized data sequence into subsequences of length 10 for the diffusion process, and train the model for approximately 600,000 iterations on a single RTX 4090. During training, the diffusion step \( t \) is uniformly sampled from 0 to 1000. During inference, an acceleration technique is applied as follows. First, \( t \) is initialized to 1000 and directly denoised to \( x_{0}^{\text{pred}} \). Subsequently, noise is added to the \( t = 1000 - 100N_i \) level and denoised again to \( x_{0}^{\text{pred}} \), where \( N_i \) is the inference step number. Thus, the entire inference process consists of 10 steps.   

For terrain generation, we model the terrain beneath each finger as a cubic spline with a trajectory length of 100. Control points are placed at intervals of 25 along the trajectory, resulting in a total of 5 control points. To simulate ramps, the height of each control point is randomized by sampling uniformly within the range of $[0,3]$ cm.

\subsection{Depth Reconstruction of VBTS}\label{exp:depth_recon}

To evaluate the performance of the tactile sensor in depth reconstruction, the sensor surface is pressed with three indenters, each with the text content “RSS”, “$\bigstar$” and “2025”. The qualitative results of the sensor output are shown in \Cref{fig:reconstruction}, which demonstrates the raw captured image from the sensor, the ground truth depth maps, predicted depth maps, and the corresponding calculated normal maps, respectively. These results demonstrate that the sensor can fully reconstruct fine surface details.

We quantify the reconstruction error using a violin plot, leveraging ground truth indentation information obtained from 3D-printed hemispherical shape indicators containing various testing indenters. We collected 215 testing configurations, each with paired sensor outputs and ground truth reprojection images. The sensor achieves a mean absolute error (L1 error) reconstruction loss of 0.35 mm, and a median loss of 0.28 mm, with 60\% of reconstruction losses below 0.3 mm. In terms of computational speed, the depth mapping process takes less than 10 ms, ensuring real-time performance for robotic applications.

\subsection{Evaluation of PP-Tac Policy on Materials and Terrains}\label{sec:comparison}
We conducted experiments to evaluate the system's ability to handle flat objects under varying conditions. The qualitative and quantitative results are shown in \Cref{fig:gallery} and \Cref{fig:comparison} respectively.
\Cref{fig:gallery} shows the typical successful grasp cases, highlighting that our hardware and PP-Tac algorithm can successfully handle flat objects placed above both the flat and uneven object surface. During the grasping process, the fingertip first contacts the material, followed by a gradual finger closure that buckles the material and creates pinchable regions. Finally, the object is pinched and lifted. 

\Cref{fig:comparison} provides quantitative analysis of the success rate with respect to the object material and the complexity of the terrain beneath. To facilitate this analysis, we conducted experiments using four flat objects in daily life: paper, plastic bag, cloth, and kraft paper bag, each of which presents unique challenges. The paper is extremely flat with no detectable hold points. Plastic bags, commonly encountered in daily life, are difficult to locate using conventional visual pipelines because of their transparency. The cloth is thick and highly deformable, while the kraft paper bags are stiff and have a multilayered structure.
To assess the system's robustness, we also varied the terrain beneath the objects. The four types of terrain used include: a flat plane, a slope (10 degrees), a plane with a 2 cm thick book randomly placed on it, and an uneven terrain with random curvatures. The terrain shapes are shown in \Cref{fig:comparison}. 

For statistical significance, we performed 20 grasping attempts for each combination of terrain and object. From results in \Cref{fig:comparison}, cloth and plastic bags are relatively easy to grasp due to their low stiffness, which allows them to buckle more easily under force. In contrast, paper and kraft paper bags are stiffer and resist buckling, leading to lower success rates.

The terrain beneath the object also significantly impacts grasp success. On flat terrains, such as a plane or a tilted slope, success rates for paper, plastic bags, and cloth were relatively high. This suggests that flat surfaces usually generate consistent frictional forces essential for a successful grasp. However, this advantage diminishes for stiffer flat objects, such as kraft paper bags. These stiff flat objects usually lack of initial buckling when placed on a flat surface, making it more challenging to form reliable grasp points afterward.

For uneven surfaces, the success rates varied according to the shape of the terrain. When a book was placed underneath the flat object, all objects maintained high success rates. These results can be attributed to the edge of the book and the partial void space created beneath the material, which made it easier for the materials to buckle and separate with the terrain. In contrast, when the terrain was highly irregular, the success rate dropped for all objects. This is likely due to the challenges added to our force controllers, which increased the likelihood of the fingers slipping away from the material.

\begin{table}[ht!]
    \centering
    \caption{\textbf{Experimental results for varying paper quantities:} The system’s performance was evaluated on paper materials with different buckling strengths, achieved by bonding 1, 3, 5, and 7 layers of paper with adhesive. For each configuration, 20 trials of grasps were conducted. The average number of slip events detected (No. Slip) and the final success rate (Succ. Rate) were recorded.}
    \begin{tabular}{ccc}
        \hline\hline
        \textbf{Paper Layers} & \textbf{No. Slip} & \textbf{Succ. Rate (\%)} \\ \hline
        1                      & 0.2             & 90                 \\ \hline
        3                      & 2.9             & 75                 \\ \hline
        5                      & 13.3            & 30                 \\ \hline
        7                      & 18.2            & 5                  \\ \hline\hline
    \end{tabular}
    \label{tab:paper_num}
\end{table}

\subsection{Comparison with Other System Configurations}\label{sec:exp:other_conf}
To assess whether PP-Tac's system setup leveraging dexterous hand and tactile sensors can offer advantages, systematic comparisons with other robot configurations were conducted. Here, we constructed three baselines. To ensure fairness, each trial allowed only one grasp attempt.

\begin{itemize}
\item Bi-finger grippers controlled via human teleoperation with a camera mounted on the wrist to provide an egocentric view which can mimic the vision-based method \cite{chi2023diffusion}. This baseline can demonstrate the effectiveness of our hardware design. 

\item Open-loop control without tactile feedback: we pre-generated trajectories using the ground truth shape of the terrain and then replayed these trajectories rather than using the PP-Tac policy. Note that this trajectory-replay setting is unattainable in scenarios with high variations, such as the book setting and the complex terrain scenario in which the terrain shape is unknown.

\item ``Model based force tracking'': 
due to the challenges outlined in \cref{sec:problem_statement}, we employ the wrist trajectory generated by PP-Tac while actively controlling only the fingertips through real-time tactile feedback.
\end{itemize}

The evaluation results in \Cref{fig:comparison} show that the PP-Tac pipeline outperforms all baselines. 
We observed that the teleoperation baseline using a gripper achieved some successful cases in grasping cloth and plastic bags, albeit with lower performance than PP-Tac. This is due to the ease of detecting the initial grasp point on these soft materials through human perception, and combined with human intelligence enabling grasp adjustments through visual feedback. However, for stiffer materials like paper and kraft paper, the bi-finger gripper failed completely. Therefore, we conclude that the PP-Tac pipeline is the most suitable configuration for handling flat objects.
The open-loop baseline achieved a lower success rate compared to PP-Tac. The suboptimal performance primarily stems from control error. As mentioned in \cite{shaw2023leaphandlowcostefficient}, Allegro Hand \cite{AllegroHand} exhibits joint angle errors exceeding 0.1 radians, which will be further accumulated across the kinematic chain. These errors critically degrade performance in precision-sensitive tasks such as paper picking, highlighting the necessity of tactile feedback for robust control.
While the "Model-based force tracking" achieves satisfactory performance in structured terrains by leveraging wrist trajectories generated by PP-Tac, its effectiveness becomes limited when confronted with irregular or complex terrains. This underscores the need for enhanced adaptability in unstructured environments.

\subsection{Ablation Studies} \label{sec:ablation}
\subsubsection{Influence of Material Stiffness} We found that the material's stiffness (represented by its thickness), significantly influences the task's success rate. To demonstrate this effect, we created flat objects by stacking paper pages bonded with adhesive. The experimental results are shown in ~\cref{tab:paper_num}. As the number of paper pages increased, the grasp success rate decreased significantly. Additionally, the increase in material stiffness also led to a higher number of detected slips.

\subsubsection{Influence of Data Disturbance} We emphasize the importance of the data disturbance technique for domain randomization (introduced in \cref{subsec:DiffusionPolicy}). To quantify its impact, we conducted ablation studies comparing grasp performance before and after adding four types of disturbances to the prefix motion \(x^{prefix}\). Experimental results demonstrate that this technique significantly enhances performance. As shown in the ``Non-disturbance'' baseline in \Cref{sec:comparison}, removing data disturbance led to a notable performance drop across all experiments, often resulting in complete failure when grasping stiff objects, such as kraft paper bags. This underscores the improved generalization and higher grasp success rates enabled by domain randomization. However, a drawback of this technique is the increased training time, requiring approximately 400,000 additional iterations to achieve the same loss as training without data disturbance.

\section{Limitations}
We have observed the following limitations in our system. One limitation is determining the initial force (sensor's target deformation depth) required for successful grasping. While our algorithm can adaptively adjust the force magnitude online, an appropriate initial value must still be manually set, which remains an empirical parameter-tuning process. If the initial value is too small, the grasp is more likely to fail due to the additional time and finger sliding distance needed for adaptation to a reasonable value. Conversely, if the initial value is too large, excessive friction may exceed the load capacity of the hand motors. In addition to the initial value, the adaptive algorithm for adjusting force also has room for improvement, particularly with highly stiff materials such as kraft paper bags on non-flat surfaces.

\section{Conclusions}\label{sec:conculsion}
This paper presents PP-Tac, a coordinated hand-arm system designed to manipulate thin, flat objects such as paper and fabric. The system is equipped with a multi-fingered, vision-based tactile sensor that is easy to fabricate and deploy on the hand’s fingertips. The sensor can detect contact on its curved surfaces, enabling the system to measure force and friction during contact. This capability helps minimize slip and increases the likelihood of material deformation when handling flat materials. Based on this hand design, the grasping motion is planned using a data-driven approach. We developed an efficient synthesis algorithm to generate sliding trajectories across various terrain shapes and sensor deformation conditions, resulting in a dataset of 500,000 trajectory samples. Using this dataset and a domain randomization technique, we trained a diffusion policy that enables adaptation to diverse terrains in real-world settings. Experimental results show that our system can successfully grasp flat objects of varying thicknesses and stiffness, achieving a success rate of 87.5\%. Additionally, the proposed policy demonstrates robustness to external disturbances and adapts well to different support terrain surfaces.

\section{Acknowledgment}

We thank Changyi Lin (Carnegie Mellon University) for insightful discussions. This work was supported in part by the National Natural Science Foundation of China (Grant No.52305007), by the State Key Laboratory of Mechanical System and Vibration (Grant No. MSV202519), and by Shanghai Frontiers Science Center of Human-centered Artificial Intelligence (ShangHAI), MoE Key Laboratory of Intelligent Perception and Human-Machine Collaboration (KLIP-HuMaCo).

    \bibliographystyle{plainnat}
    \bibliography{references}

\appendix

\subsection{Detail of Camera Calibration}\label{appendix:cameracali}

\begin{figure}[th!]
    \centering
    \includegraphics[width=\linewidth]{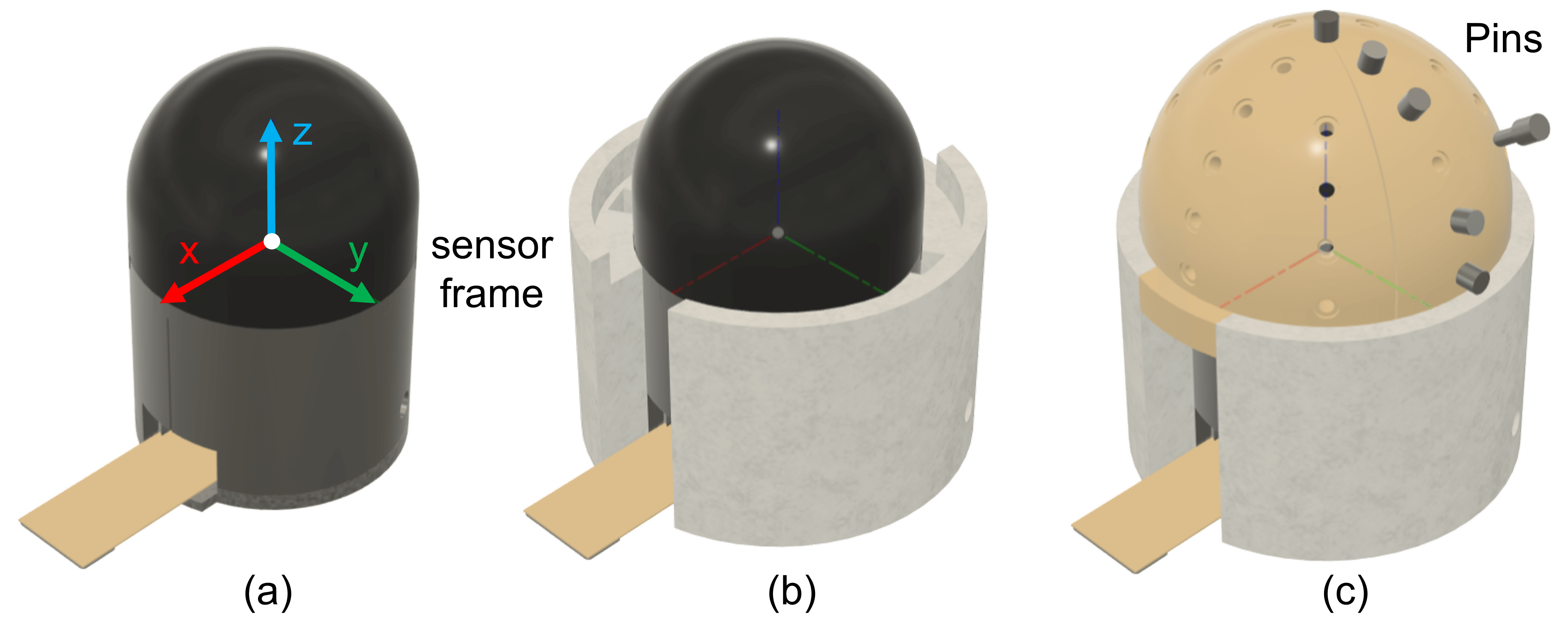}
    \caption{\textbf{Camera calibration using an indentation setup:} The sensor frame is first defined in (a). A holder is designed and 3D-printed to secure the sensor, as shown in (b). A hemispherical indicator is designed and 3D-printed to attach to the sensor holder. Pins are inserted into pre-defined holes to serve as indenters for recording contact locations in the sensor frame, as shown in (c).}
    \label{fig:cameracali}
    \vspace{-10pt}
\end{figure}

In this section, we introduce the camera calibration process as part of the overall sensor calibration. Since the tactile sensor is enclosed by an opaque, rounded membrane, conventional calibration board methods cannot be used to determine the pinhole camera's extrinsic parameters. To address this, we designed an indentation setup (as shown in ~\cref{fig:cameracali}) to capture a sufficient number of spatial points in a known sensor frame, identify their corresponding 2D-pixel coordinates in the image, and establish the mapping between the sensor frame and the image frame. First, the camera's intrinsic parameters $K$ was obtained, either from the camera manufacturer or calibrated using high-precision calibration boards~\cite{zhang2002flexible}. Next, we define a three-dimensional coordinate system, referred to as the sensor frame ($x,y,z$) with its origin at the center of the elastomer, as shown in \cref{fig:cameracali}(a). To facilitate the calibration, A custom 3D-printed holder secures the sensor (\cref{fig:cameracali}(b)), while another 3D-printed hemispherical indicator is attached to the holder's groove (\cref{fig:cameracali}(c)). Small pins with a diameter of $1.5$mm, serving as indenters, are inserted into pre-defined holes within the indicator for 28 trials. For each trail, the contact positions are recorded both in the camera image as $p_{ij}=(u_{ij}, v_{ij})$ and in the sensor frame as $P_{i,j}=(x_{ij},y_{ij},z_{ij})$, where $i$ denotes the trail index and $j$ denotes the contact point index within the trail. The contact positions in the camera image are detected by subtracting the captured image from a reference image without indentation. We use solvePnP~\cite{lepetit2009ep} to calculate the extrinsic parameters that includes rotation matrix $A$ and translation vector $b$ such that:

\begin{equation}
    p_{ij} = K[A \mid b] P_{ij}
\end{equation}

After obtaining the intrinsic and extrinsic parameters of the camera, we can project the sensor's curved surface from the sensor frame onto the image frame, obtaining the sensor surface reference projection $D$ (\cref{eq:Duv}), by which the depth value on the pixel $(u,v)$ can be queried. 

\begin{equation}
     D(u,v) = \begin{bmatrix}Z_c K^{-1}
     \begin{bmatrix}
     u\\
     v\\
     1
    \end{bmatrix}
    \end{bmatrix}_{[3, :]},
    \label{eq:Duv}
\end{equation}

where $[u\; v\; 1]^T$ and $Z_c$ are given as:
\begin{equation}
     \begin{bmatrix}
     u\\
     v\\
     1
    \end{bmatrix} = K
    \begin{bmatrix}
    \frac{(A[x,y,z]^T+b)_x}{Z_c} \\
    \frac{(A[x,y,z]^T+b)_y}{Z_c} \\
    1
    \end{bmatrix}, Z_c = (A[x,y,z]^T+b)_z.
\end{equation}

\subsection{Detail of Establish Contact}\label{appendix:InitializeState}
    
    \begin{figure}[t!]
        \centering
        \includegraphics[width=\linewidth]{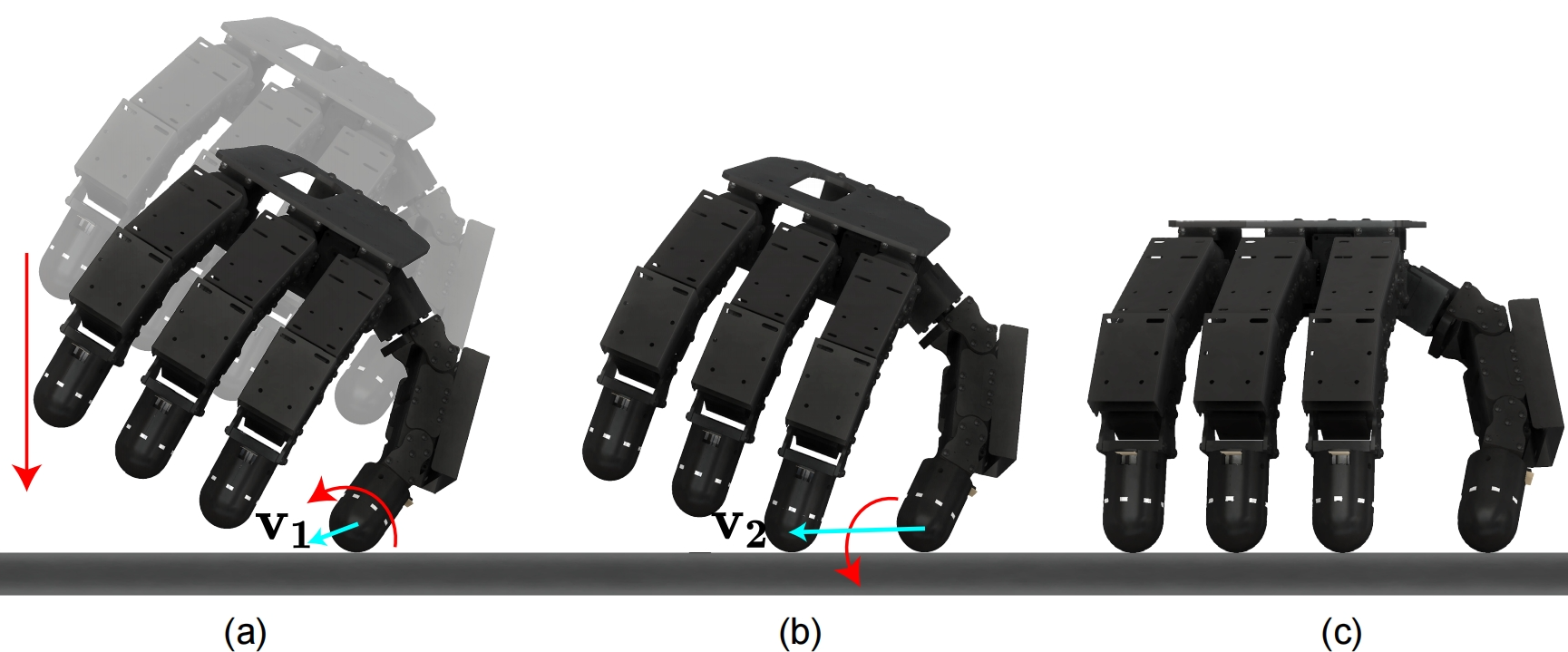}
        \caption{\textbf{Example of establishing contact:} First, the hand descends until a finger makes contact with the surface. A fixed-point rotation is performed around the contacting finger, as shown in (a). The hand then continues to rotate until a second finger makes contact, triggering a fixed-axis rotation around both contacting fingers, as shown in (b). The process is complete when three or more fingers are in contact, as shown in (c).}
        \label{fig:initialize}
        \vspace{-10pt}
    \end{figure}

In this section, we detail our approach to generate contact with a flat object using the fingertips. The goal is to control the hand to ensure that at least three fingertips are in contact with the surface. We denote the four fingertips as \(f_t\) (thumb), \(f_i\) (index), \(f_m\) (middle), and \(f_r\) (ring). The contact states are represented by two sets: \(\pmb{C}\), which includes the fingers in contact, and \(\pmb{N}\), which includes the fingers not in contact. The complete process is illustrated in \cref{fig:initialize}.

    \subsubsection{ \textbf{Establish First Contact}} Starting from status when all fingers are hovering (i.e., \(\pmb{C} = \phi  ,\pmb{N}=\left \{ f_{t},f_{i},f_{m},f_{r} \right \}\)), the hand is controlled to move downward till one finger touches the surface. For example, if the thumb touches the surface (\cref{fig:initialize}), the contact state sets are updated to \(\pmb{C} =\left \{ f_{t} \right \} ,\pmb{N}=\left \{ f_{i},f_{m},f_{r} \right \}\)
    
    \subsubsection{ \textbf{Establish Second Contact}} Once the first contact is made, the hand rotates around the first finger's contact point to create the second contact point. To achieve this, we first obtain the centroid point of the fingertip in contact (denoted as \((x_{c},y_{c},z_{c})\)),  and compute the centroid point of fingertip positions in \(\pmb{N}\) (denoted as \((x_{n},y_{n},z_{n})\)). This allows us to calculate the rotational axis as: 
\begin{equation}
    \pmb{v_{1}} = R_{z}(90^{\circ})(x_{n}-x_{c},y_{n}-y_{c},z_{n}-z_{c})^T,
\end{equation}
where \(R_{z}(90^{\circ})\) is the rotation matrix for a 90-degree rotation around the z-axis. Given $\theta,\pmb{v_{1}}$ calculated before, robot arm's target end-effector pose $^{\mathit{b}}_{\mathit{ee'}}T$ leading to such rotation can be obtained via Rodrigues' rotation formula:

    \begin{align}
    R(\theta, \pmb{v_1}) = &I + \sin(\theta) 
    \begin{bmatrix}
    0 & -{v_{1z}} & {v_{1y}} \\
    {v_{1z}} & 0 & -{v_{1x}} \\
    -{v_{1y}} & {v_{1x}} & 0
    \end{bmatrix} + \notag \\
     (&1 - \cos(\theta))
    \begin{bmatrix}
    0 & -{v_{1z}} & {v_{1y}} \\
    {v_{1z}} & 0 & -{v_{1x}} \\
    -{v_{1y}} & {v_{1x}} & 0
    \end{bmatrix}^2,
    \end{align}

The target end effector pose of the robot arm can be calculated as:

\begin{equation}
    ^{\mathit{b}}_{\mathit{ee'}}T = ^{\mathit{b}}_{\mathit{ee}}T \ ^{\mathit{ee}}_{\mathit{c}}T \, ^{\mathit{c}}_{\mathit{c'}}\hat{T} \, {^{\mathit{c'}}_{\mathit{ee'}}T},
\end{equation}

\begin{equation}
^{\mathit{c}}_{\mathit{c'}}\hat{T} = \begin{bmatrix} 
R(\theta, \pmb{v_1}) & 0 \\
0 & 1
\end{bmatrix},
\end{equation}

where \(b\) denotes the base of the robot arm, \({ee}\) and \({ee'}\) represent the end effector before and after the movement, and \(c\) and \(c'\) represent the positions \((x_c, y_c, z_c)\) before and after the rotation. The robot arm is then controlled to gradually increase \(\theta\) until the second fingertip contacts the object surface. Once this occurs, we update the contact states to \(\pmb{C} = \{ f_t, f_i \}\) and \(\pmb{N} = \{ f_m, f_r \}\).

    %

    %

\subsubsection{ \textbf{Establishing Third Contact}}  
In this step, the hand rotates around an axis defined by the first and second contact points until the third fingertip makes contact. For instance, if the thumb and index finger make contact, the rotation axis is \(\pmb{v_2} = \overrightarrow{f_t f_i}\). The arm's target end-effector pose for this rotation is:

\begin{equation}
   ^{\mathit{b}}_{\mathit{ee''}}T = ^{\mathit{b}}_{\mathit{ee'}}T \ ^{\mathit{ee'}}_{\mathit{c'}}T \, ^{\mathit{c'}}_{\mathit{c''}}\hat{T} \, {^{\mathit{c''}}_{\mathit{ee''}}T}, 
\end{equation}

\begin{equation}
^{\mathit{c'}}_{\mathit{c''}}\hat{T} = \begin{bmatrix} 
R(\theta', \pmb{v_2}) & 0 \\
0 & 1
\end{bmatrix},
\end{equation}

where $c''$ and $ee''$ are $c'$ and $ee'$ after rotation specified by $\pmb{v_2}$. During execution, the angle \(\theta'\) is gradually increased until a new fingertip contacts the surface, achieving the desired target end-effector pose \(^{\mathit{b}}_{\mathit{ee'}}T\). Note that these steps may not always be required. In some cases, we observe that the third finger may already be in the contact state when we attempt to establish contact with the second finger.

   

\subsection{List of Symbols}\label{appendix:List of Symbols}
The definition of symbols can be found in ~\cref{tab:symbols}.

\begin{table}[ht!]
\centering
\caption{Summary of symbols and notations. } 
\label{tab:symbols}
\begin{tabular}{cp{0.8\linewidth}}
\toprule
\textbf{Symbols}              & \textbf{Descriptions} \\ \midrule
\(u,v\)                       & Pixel coordinates in \ac{vbts}. \\
\(X_c, Y_c, Z_c\)              & Camera coordinates in \ac{vbts}. \\
\(x, y, z\)                    & Sensor coordinates in \ac{vbts}. \\
\(K\)                        & The intrinsic parameters of the camera in \ac{vbts}. \\
\(A, b\)                     & The extrinsic parameters of the camera in \ac{vbts}. \\
\(D\)                        & Sensor surface reference projection in \ac{vbts}. \\
\(M\)                        & Depth mapping function in \ac{vbts}. \\
\(\pmb{q}\)                  & Rotation angle of controllable hand joints. \\ 
\(\dot{\pmb{q}}\)            & Angular velocity of controllable hand joints. \\ 
\(\pmb{p}\)                & Positional coordinate of hand joints in arm's base axis. \\ 
\(\dot{\pmb{p}}\)          & Linear velocity of hand joints in arm's base axis. \\ 
\(R\)                        &  Wrist's (end effector of arm) 6D rotation. \\ 
\(\Omega\)                   & Angular velocity of hand pose. \\ 
\(p_{wrist}\)             & Wrist (end-effector of arm)’s height along arm’s \(z\)-axis. \\ 
\(\dot{p}_{wrist}\)       & Linear velocity of \(\pmb{p}_{ee}\). \\ 
\(\pmb{d}_{tac}\)            & The deformation depth readings from four fingertip tactile sensors. \\ 
\(\bar{\pmb{d}}_{tac}\)      & The target deformation depth. \\ 
\(\mathcal{D}\)              & State variable's dimension. \\ 
\(\gamma\)                   & Hand joint angles \(\pmb{q}^{1:N_{data}}\), wrist's (end effector of arm) 6D rotation \(R^{1:N_{data}}\) and wrist's translation along z-axis \(p_{ee}^{1:N_{data}}\) for overall trajectory. \\ 
\(N_{data}\)                 & Length of synthesis motion sequence. \\ 
\(N_{pred}\)                 & Length of predicted actions. \\ 
\(x^{pred}\)                 & Future motion predicted by PP-Tac policy. \\ 
\(N_{prefix}\)               & Length of historical actions. \\ 
\(x^{prefix}\)               & The historical action sequence. \\
\(t\)                        & Diffusion step. \\ \bottomrule
\end{tabular}
\end{table}
\end{document}